\documentclass[sigconf]{acmart}

\setcopyright{acmlicensed}
\copyrightyear{2019}
\acmYear{2019}
\acmConference[MM '19]{Proceedings of the 27th ACM International Conference on Multimedia}{October 21--25, 2019}{Nice, France}
\acmBooktitle{Proceedings of the 27th ACM International Conference on Multimedia (MM '19), October 21--25, 2019, Nice, France}
\acmPrice{15.00}
\acmDOI{10.1145/3343031.3351066}
\acmISBN{978-1-4503-6889-6/19/10}

\usepackage{breakurl}
\usepackage{makecell}
\usepackage{tabularx}
\usepackage{multirow}

\usepackage{balance}
\def\BibTeX{{\rm B\kern-.05em{\sc i\kern-.025em b}\kern-.08emT\kern-.1667em\lower.7ex\hbox{E}\kern-.125emX}}

\begin{document}

\fancyhead{}
  % do not delete this code.

% The "title" command has an optional parameter, allowing the author to define a "short title" to be used in page headers.
\title{Towards Automatic Face-to-Face Translation}
\author{Prajwal K R}
\authornote{Both authors contributed equally to this research.}
\email{prajwal.k@research.iiit.ac.in}
\affiliation{%
  \institution{IIIT Hyderabad}
  %\city{Hyderabad}
}

\author{Rudrabha Mukhopadhyay}
\authornotemark[1]
\email{radrabha.m@research.iiit.ac.in}
\affiliation{%
  \institution{IIIT Hyderabad}
  %\city{Hyderabad}
}

\author{Jerin Philip}
\email{jerin.philip@research.iiit.ac.in}
\affiliation{%
  \institution{IIIT Hyderabad}
  %\city{Hyderabad}
}
 
\author{Abhishek Jha}
\email{abhishek.jha@research.iiit.ac.in}
\affiliation{%
  \institution{IIIT Hyderabad}
  %\city{Hyderabad}
}

\author{Vinay Namboodiri}
\email{vinaypn@iitk.ac.in}
\affiliation{%
  \institution{IIT Kanpur}
  %\city{Kanpur}
}

\author{C. V. Jawahar}
\email{jawahar@iiit.ac.in}
\affiliation{%
  \institution{IIIT Hyderabad}
  %\city{Hyderabad}
}

% \input{author-standard.tex}

%
% By default, the full list of authors will be used in the page headers. Often, this list is too long, and will overlap
% other information printed in the page headers. This command allows the author to define a more concise list
% of authors' names for this purpose.
\renewcommand{\shortauthors}{Prajwal and Rudrabha, et al.}

%
% The abstract is a short summary of the work to be presented in the article.
\begin{abstract}
In light of the recent breakthroughs in automatic machine translation systems, we propose a novel approach that we term as "Face-to-Face Translation". As today's digital communication becomes increasingly visual, we argue that there is a need for systems that can automatically translate a video of a person speaking in language A into a target language B with realistic lip synchronization. In this work, we create an automatic pipeline for this problem and demonstrate its impact in multiple real-world applications. First, we build a working speech-to-speech translation system by bringing together multiple existing modules from speech and language. We then move towards "Face-to-Face Translation" by incorporating a novel visual module, LipGAN for generating realistic talking faces from the translated audio. Quantitative evaluation of LipGAN on the standard LRW test set shows that it significantly outperforms existing approaches across all standard metrics. We also subject our Face-to-Face Translation pipeline, to multiple human evaluations and show that it can significantly improve the overall user experience for consuming and interacting with multimodal content across languages. Code, models and demo video are made publicly available.
\end{abstract}



%
% Keywords. The author(s) should pick words that accurately describe the work being
% presented. Separate the keywords with commas.

%
% A "teaser" image appears between the author and affiliation information and the body 
% of the document, and typically spans the page. 
\begin{teaserfigure}
  \includegraphics[width=\textwidth]{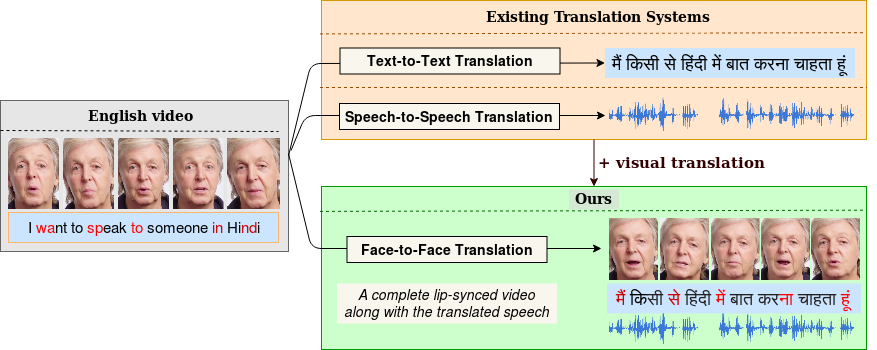}
  \caption{In light of the increasing amount of audio-visual content in our digital communication, we examine the extent to which current translation systems handle the different modalities in such media. We extend the existing systems that can only provide textual transcripts or translated speech for talking face videos to also translate the visual modality i.e. lip and mouth movements. Consequently, our proposed pipeline produces fully translated talking face videos with corresponding lip synchronization.}
  \label{fig:teaser}
\end{teaserfigure}

%
% This command processes the author and affiliation and title information and builds
% the first part of the formatted document.
\maketitle

\section{Introduction}
Communicating effectively across language barriers has always been a major aspiration for humans all over the world. In recent years, there has been tremendous progress by the research community towards this goal. Neural Machine Translation (NMT) systems have become increasingly competent\cite{wu2016google,bahdanau2014neural,vaswani2017attention} in automatically translating foreign languages without the need for a human in the loop. The success of the recent NMT systems not only impacts plain text-to-text translation but also plays a pivotal role in speech-to-speech translation systems. The latter problem is also of great interest because a large part of our communication with others is oral. By cascading speech recognition, neural machine translation and speech synthesis modules, current systems can generate a translated speech output for a given source speech input\cite{skype,federmann2016microsoft}. In this work, we argue that it is possible to extend this line of research further with a visual module that can greatly broaden the scope and enhance the user experience of existing speech translation systems.

The motivation to incorporate a visual module into a translation system arises from the fact that the majority of the information stream today, is increasingly becoming audio-visual. YouTube, the world's largest online video sharing platform generates 300 hours of video content every minute\footnote{https://merchdope.com/youtube-stats/}. The meteoric rise of video conferencing~\cite{videocalling} also exemplifies the preference for rich audio-visual communication. Existing systems can only translate such audio-visual content at a speech-to-speech level and hence possess some major limitations. Firstly, the translated voice sounds very different from the original speaker's voice. But, more importantly, the generated speech when directly overlaid on the original video produces unsynchronized lip movements with respect to the speech, leading to poor user experience. Thus, we build upon the speech-to-speech translation systems and propose a pipeline that can take a video of a person speaking in a source language and output a video of the same speaker speaking in a target language such that the voice style and lip movements justifies the target language. By doing so, the translation system becomes holistic, and as shown by our human evaluations in this paper, significantly improves the user experience in creating and consuming translated audio-visual content.  

Our pipeline is made up of five modules. In the scope of this paper, we work with two widely spoken languages: English and Hindi. For speech to speech translation, we use an automatic speech recognizer\cite{amodei2016deep} to transcribe text from the original speech in language L$_A$. We adapt state-of-the-art neural machine translation and text-to-speech models\cite{vaswani2017attention,ping2017deep} to work for Indian languages and generate translated speech in language L$_B$. We also personalize the voice\cite{kaneko2017parallel} generated by the TTS model to closely match the voice of the target speaker. 

Finally, to generate talking faces conditioned on the translated speech, we design a novel generative adversarial network, ~\textit{LipGAN} in which we employ an adversary that measures the extent of lip synchronization in the frames generated by the generator. Furthermore, our system is capable of handling faces in random poses without the need for realignment to a template pose. Our intuitive approach yields realistic talking face videos from any audio with no dependence on language. We achieve state-of-the-art scores on the LRW test set across all quantitative metrics. Using our complete pipeline, we show a proof-of-concept on multiple applications and also propose future directions in this novel research problem. Different resources for this work along with demo videos are available publicly\footnote{http://cvit.iiit.ac.in/research/projects/cvit-projects/facetoface-translation}. In summary, our contributions are as follows:
\begin{enumerate}
    \item For the first time, we design and train an automatic pipeline for the novel problem of face-to-face translation. Our system can automatically translate a talking face of a person into a given target language, with realistic lip synchronization.
    \item We propose a novel model, LipGAN, for generating realistic talking faces conditioned on audio in any language. Our model outperforms existing works in both quantitative and human-based evaluation. 
    \item In the process of creating a face-to-face translation pipeline, we also achieve state of the art neural machine translation results in the Hindi-English language pair by incorporating recent advancements in the area.
\end{enumerate}

The rest of the paper is organized as follows: In section \ref{section:background}, we survey the recent developments in speech, vision and language research which enable our work. Following this, the adaptation of the existing methods in speech and language to our problem setting is described in section \ref{section:adaptations}. Section \ref{section:lipgan} explains in detail our novel contributions to bring improvements in lip synchronization. We produce a few applications deriving from our work in Section \ref{section:applications} and conclude our findings in Section \ref{section:conclusion}. 

\section{Background}

\label{section:background}
Given a video of a speaker speaking in language L$_A$, our aim is to generate a lip-synchronized video of the speaker speaking in language L$_B$. Our system brings together multiple modules from speech, vision, and language to achieve face to face translation for the first time. 

\subsection{Automatic Speech Recognition}
We make use of recent works on Automatic Speech Recognition (ASR)\cite{amodei2016deep} to convert the speech of the source language L$_A$ into the corresponding text. Speech recognition for English has been extensively investigated, owing to the existence of large open-source speech recognition datasets\cite{panayotov2015librispeech,rousseau2012ted} and trained models\cite{amodei2016deep}. We employ the DeepSpeech 2 model to perform English speech recognition in this work.

%Another option worth considering was whether to directly obtain the translated text from the source speech. Recent works\cite{weiss2017sequence} show the possibility of training such a system on Spanish-English parallel data. However, the lack of large, readily available parallel corpora for our chosen languages and the flexibility that comes with splitting our entire system into independent modules convinced us against the speech to translated text approach.
\begin{figure*}[h]
  \includegraphics[width=\linewidth]{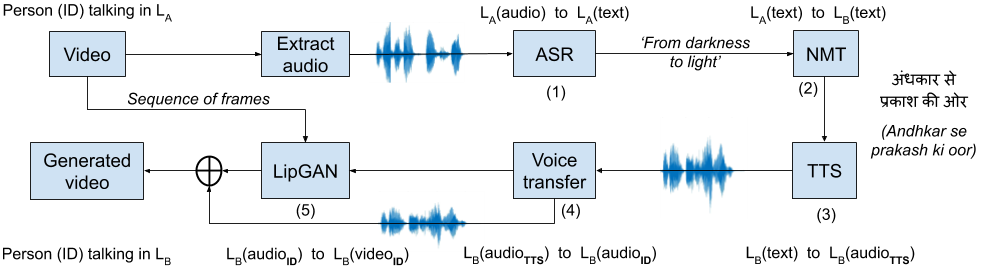}
  \caption{Block diagram of the overall pipeline of our network. In our case, L$_A$ is English and L$_B$ is Hindi. We decompose our problem into: (1) recognize speech in the source language L$_A$, (2) translate the recognized text in L$_A$ to a target language L$_B$, (3) synthesize speech from the translated text (5) generate realistic talking faces in language L$_B$ from the synthesized speech. Additionally, to obtain personalized speech for a speaker, we employ a Voice transfer module (4).}
  \label{fig:overallpipeline}
\end{figure*}

\subsection{Neural Machine Translation}
NMT is often modelled as a sequence to sequence problem which was first introduced with neural networks in ~\citet{sutskever2014sequence}. % 
% The paper used an encoder-decoder based architecture made up of RNNs. Further improvements were brought about by ~\citet{bahdanau2014neural} introducing attention. ~\citet{luong2015effective} used attention based frameworks which also comprised of RNN based architectures but the decoders in this case used context vectors from encoders to predict the final output. ~\citet{gehring2017convolutional} introduced fully convolutional sequence to sequence learning in, which proved to be a better performer when compared to other sequence to sequence networks. 
Further improvements were brought about with attention mechanisms by ~\citet{bahdanau2014neural} and ~\citet{luong2015effective}. More recently, ~\citet{vaswani2017attention} introduced transformer network which relies only on an attention mechanism to draw global dependencies between input and output. The transformer network outperforms its predecessors by a healthy margin and hence we decided to adopt this into our pipeline. It has also been observed in works like \citet{johnson2017google} that training multilingual translation systems also improve the performance especially for low resource languages. Thus, in this work, we also follow a similar path where we use state of the art architectures extended to multilingual learning setups. 

\subsection{Text to Speech}
There has been a lot of work in the area of text-to-speech (TTS) synthesis, starting with the most commonly used HMM-based models\cite{zen2009statistical}. These models can be trained with lesser data to produce fairly intelligible speech, but fail to capture aspects like prosody that is evident in natural speech. Recently, researchers have achieved natural TTS by training neural network-based architectures~\cite{shen2018natural,ping2017deep} to map character sequences to mel-spectrograms. We adopt this approach, and train Deep Voice 3\cite{ping2017deep} based models to achieve high-quality text-to-speech synthesis in our target language L$_B$. Our implementation of DeepVoice 3 also makes use of a recent work on guided-attention~\cite{tachibana2018efficiently} allowing it to achieve high-quality alignment and faster convergence. 

\subsection{Voice Transfer in Audio}
Multiple recent works~\cite{ping2017deep,arik2018neural} make use of multi-speaker TTS models to generate voice conditioned on speaker embeddings. While these systems offer the advantage of being able to generate novel TTS voice samples given a few seconds of reference audio, the quality of TTS is inferior~\cite{ping2017deep} compared to single-speaker TTS models. In our system, we employ another recent work\cite{kaneko2017parallel} that uses a CycleGAN architecture to achieve good voice transfer between two human speakers with no loss in linguistic features. We train this model to perform a cross-language transfer of a synthetic TTS voice to a natural target speaker voice. We evaluate our models and show that by using just about ten minutes of a target speaker's audio samples, we can emulate the speaker's voice and significantly improve the experience of a listener.

\subsection{Talking Face Synthesis from Audio}
Lip synthesis from a given audio track is a fairly long-standing problem, first introduced in the seminal work of ~\citet{bregler1997video}. However, realistic lip synthesis in unconstrained real-life environments was only made possible by a few recent works~\cite{suwajanakorn2017synthesizing,kumar2017obamanet}. Typically, these networks predicted the lip landmarks conditioned on the audio spectrogram in a time window. However, it is important to highlight that these networks fail to generalize to unseen target speakers and unseen audio. A recent work by ~\citet{chung2017you} treated this problem as learning a phoneme-to-viseme mapping and achieved generic lip synthesis. This leads them to use a simple fully convolutional encoder-decoder model. Even more recently, a different solution to the problem was proposed by ~\citet{zhou2018talking}, in which they use audio-visual speech recognition as a probe task for associating audio-visual representations, and then employ adversarial learning to disentangle the subject-related and speech-related information inside them. However, we observed two major limitations in their work. Firstly, to train using audio-visual speech recognition, they use $500$ English word-level labels for the corresponding spoken audio. We observed that this makes their approach language-dependent. It also becomes hard to reproduce this model for other languages as collecting large video datasets with careful word-level annotated transcripts in various languages is infeasible. Our approach is a fully self-supervised approach that learns a phoneme-viseme mapping, making it language independent. Secondly, we observe that their adversarial networks are not conditioned on the corresponding input audio. As a result, their adversarial training setup does not directly optimize for improved lip-sync conditioned on audio. In contrast, our LipGAN directly optimizes for improved lip-sync by employing an adversarial network that measures the extent of lip-sync between the frames generated by the generator and the corresponding audio sample. Additionally, both ~\citet{zhou2018talking} and ~\citet{chung2017you} normalize the pose of the input faces to a canonical pose, thus making it difficult to blend the generated faces in the original input video. Proposed LipGAN tackles this problem by providing additional information about the pose of the target face as an input to the model thus making the final blending of the generated face in the target video fairly straightforward.

\section{Speech-to-Speech Translation}
\label{section:adaptations}

In the previous section, we surveyed the possibility of using state of the art models in speech and language to suit our problem setting. There are not many existing systems reported for speech recognition, machine translation and speech synthesis available for Indian languages. In this section, we describe the current state of the art architectures we use for text and speech, and how we adapt them to our data. 

%We decompose the problem into: (1) recognize speech in the source language L$_A$, (2) translate the recognized text in L$_A$ to a target language L$_B$, (3) synthesize speech from translated text (4) generate realistic talking faces in language L$_B$ from the synthesized speech. Additionally, to obtain personalized speech for a speaker, we employ a style transfer module. 
% We describe their architectures and their training methodologies in detail in this section.

\subsection{Recognizing speech in source language L$_A$}
We use publicly available state-of-the-art ASR systems for generating text in language L$_A$. A publicly available pre-trained model using Deep Speech 2 is used for speech recognition in English. This model was trained on LibriSpeech dataset and achieves WER\% of 5.22\% on the LibriSpeech test set. Once we have text, recognized in a source language, we translate it into a target language using an NMT model, which we discuss next.

\subsection{Translating to target language L$_B$}
%The transformer architecture ~\cite{vaswani2017attention} has enabled state of the art results in several tasks at the time of writing this paper, including machine translation. Thus, in our experiments, 
We use the re-implementation of Transformer-Base~\cite{vaswani2017attention} available in \texttt{fairseq-py}\footnote{\url{https://github.com/pytorch/fairseq}}. The language pairs we attempt our problem on contains a low resource language, Hindi. To create a {\sc nmt} system which works well for Hindi as well as English, we resort to training a multiway model to maximize learning\cite{neubig2018rapid,aharoni2019massively}. We closely follow \citet{johnson2017google} in training a multi-way model whose parameters are shared across all seven languages - Hindi, English, Telugu, Malayalam, Tamil, Telugu, Urdu. Details of the translation system has been reported in \cite{philip2019baseline}.
%and a control token in the input sequence is used to switch language directions while translating. We use only one control  token which switches the target language, and let the  network infer the source language, as suggested in ~\citet{johnson2017google}. This additionally makes the network robust to code mixed content. 
\begin{table}[h]
    \centering
    \begin{tabular}{lrr}
    \toprule
        Direction  & our-BLEU  & Online-G \\ \midrule
        Hindi to English &  22.62 & 19.58  \\
        English to  Hindi &  20.17 &  17.87 \\
         \bottomrule
    \end{tabular}
    \caption{NMT Evaluation Scores.}
    \label{tab:translation_results}
    \vspace{-0.7cm}
\end{table}
In Table \ref{tab:translation_results}, we report evaluation metrics for language directions which are within the scope of this paper. We indicate the size of training data used and the evaluated scores using the widely used Bilingual Evaluation Under Study (BLEU) obtained on the test split of IIT-Bombay Hindi-English Parallel Corpus~\cite{kunchukuttan2018iit}. We compare against Google Translate\footnote{compared in the first week of April, 2019} in this test set, which is indicated in Table \ref{tab:translation_results} as Online-G. We achieve an increase of ~3 BLEU points on the test set compared to Google Translate.

Next, we describe our methods of generating speech from the target text in L$_B$, obtained after translating source text in language L$_A$.

\subsection{Generating Speech in language L$_B$}
For our Hindi text-to-speech model, we adapt a re-implementation of the DeepVoice 3 model proposed by ~\citet{ping2017deep}. Due to the lack of publicly available large scale dataset for Hindi, we curate a dataset similar to LJSpeech by recording Hindi sentences from crawled news articles.

We adopt the {\sc nyanko-build}~\footnote{https://github.com/r9y9/deepvoice3\_pytorch} implementation of DeepVoice 3 to train our Hindi TTS model. We trained on about $10,000$ audio-text pairs and evaluated on $100$ unseen test sentences. Griffin-Lim algorithm~\cite{griffin1984signal} was used to generate waveforms from the spectrograms produced by our model. We evaluate this model by conducting a user study with $25$ participants using our unseen test set. The average Mean Opinion Scores (MOS) scores with 95\% confidence intervals are reported in Table \ref{tab:mostts}. In the next section, we describe how we can modify the voice of the TTS model to a given target speaker.

\begin{table}[h]
  \begin{tabular}{lc}
    \toprule
    Sample Type & MOS\\
    \midrule
    DeepVoice 3 Hindi & 3.56\\
    Ground truth Hindi & 4.78\\
  \bottomrule
  \end{tabular}
  \caption{The MOS for our Hindi TTS is comparable to the same architecture trained on the LJSpeech English TTS dataset.}
  \label{tab:mostts}
  \vspace{-0.7cm}
\end{table}

\subsection{Personalizing speaker voice}
Voice of a speaker is one of the key elements of her acoustic identity. As our TTS model only generates audio samples in a single voice, we personalize this voice to match the voice of different target speakers. As collecting parallel training data for the same speaker across languages is infeasible, we adopt the CycleGAN architecture~\cite{kaneko2017parallel} to work around this problem.

For a given speaker we collect about $10$ minutes of audio clips, which can be easily obtained as we need only a non-parallel dataset. Using our trained TTS model, we generate $5000$ samples amounting to about 3 hours worth of synthetic TTS speech. For each speaker, we train a CycleGAN for about $50K$ iterations with a batch size of $16$. The other hyperparameters are the same as used in ~\citet{kaneko2017parallel}. During inference, given a TTS generated audio sample, the model preserves the linguistic features and generates speech in the voice of the speaker it was trained on.

\begin{table}[h]
  \begin{tabular}{l|cc|cc}
    \hline
    Speaker & Quality & Similarity & MOS & No Transfer\\
    \hline
    Modi & 4.21 & 3.56 & \textbf{3.89} & 1.85\\
    Andrew Ng & 3.45 & 4.1 & \textbf{3.78} & 1.91 \\
    Obama & 3.64 & 2.9 & \textbf{3.27} & 1.61\\
  \hline
\end{tabular}
\caption{MOS scores for Voice Transfer of Hindi TTS on various target speakers. Using the CycleGAN approach, we are able to consistently achieve reasonable cross-language voice transfer from the TTS generated voice to a given speaker.}
\label{tab:voicemos}
\vspace{-0.7cm}
\end{table}

We evaluate our Voice Transfer models in a similar fashion to ~\citet{kaneko2017parallel}, with the help of $30$ participants. We use 20 generated TTS samples each of which are voice transferred across $5$ famous personalities. Table \ref{tab:voicemos} reports the result of this study. In the next section, we describe how we generate realistic talking face videos.

\section{Talking face generation}
\label{section:lipgan}
Given a face image $I$ containing a subject identity and a speech $A$ divided into a sequence of speech segments $\{A_1, A_2,...A_k\}$ , we would like to design a model $G$, that generates a sequence of frames $\{S_1, S_2,...S_k\}$ that contains the face speaking the audio $A$ with proper lip synchronization. Additionally, the model must work for unseen languages and faces during inference. As collecting annotated data for various languages is tedious, the model must also be able to learn in a self-supervised fashion. Table \ref{tab:lip_comp} compares our model against recent state-of-the-art approaches for talking face generation. 

\begin{table}[h]
    \centering
    {\footnotesize
    \begin{tabularx}{\linewidth}{l|X|X|X|X}
    \hline
        Method & Works for any face? & Cross -language & No manual labeled data & Smooth blending into target video \\ \hline
        ~\citet{suwajanakorn2017synthesizing} & $\times$ & $\times$ & \checkmark & \checkmark \\
        ~\citet{kumar2017obamanet} & $\times$ & $\times$ & $\times$ & \checkmark \\
        ~\citet{zhou2018talking} & \checkmark & $\times$ & $\times$ & $\times$ \\
        ~\citet{chung2017you} & \checkmark & \checkmark & \checkmark & $\times$ \\
        ~\textbf{LipGAN (Ours)} & \checkmark & \checkmark & \checkmark & \checkmark \\
         \bottomrule
    \end{tabularx}
    }
    \caption{Comparison of recent works on talking face synthesis against our LipGAN model. Ours is the first model that that generates ~\textit{realistic} in-the-wild talking face videos across languages without the need for any manually labeled data.}
    \label{tab:lip_comp}
    \vspace{-1cm}
\end{table}

\subsection{Model Formulation} 
\begin{figure*}[h]
  \includegraphics[width=\textwidth]{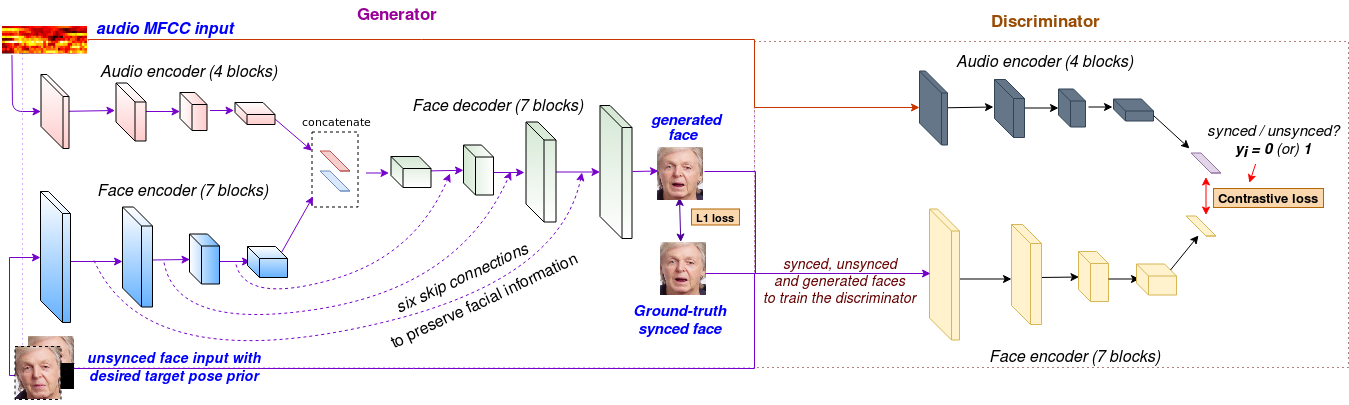}  \caption{We train our LipGAN network in an intuitive GAN setup. The generator generates face images conditioned on the audio input. The discriminator checks whether the generated frame and the input audio are in sync. Note that while training the discriminator, we also feed extra ground-truth synced / unsynced samples to ensure that the discriminator learns to specifically check for superior lip-sync and not just the image quality.}
  \label{fig:arch}
  \vspace{-0.3cm}
\end{figure*}

We formulate our talking face synthesis problem as ``learning to synthesize by testing for synchronization". Concretely, our setup contains two networks, a generator $G$ that generates faces by conditioning on audio inputs and a discriminator $D$ that tests whether the generated face and the input audio are in sync. By training these networks together in an adversarial fashion, the generator $G$ learns to create photo-realistic faces that are accurately in sync with the given input audio. The setup is illustrated in Figure \ref{fig:arch}.

\subsection{Generator network} The generator network is a modification of ~\citet{chung2017you} and contains three branches: $(i)$ Face encoder, $(ii)$ Audio encoder and a $(iii)$ Face Decoder.

\subsubsection{The Face Encoder.} We design our face encoder a bit differently from ~\citet{chung2017you}. We observe that during the training process of the generator in ~\cite{chung2017you}, a face image of random pose and its corresponding audio segment is given as input and the generator is expected to morph the lip shape. However, the ground-truth face image used to compute the reconstruction loss is of a completely different pose, and as a result, the generator is expected to change the pose of the input image without any prior information. To mitigate this, along with the random identity face image $I$, we also provide the desired pose information of the ground-truth as input to the face encoder. We mask the lower half of the ground truth face image and concatenate it channel-wise with $I$. The masked ground truth image provides the network with information about the target pose while ensuring that the network never gets any information about the ground truth lip shape. Thus our final input to the face encoder is a $H$x$H$x$6$ image. The encoder consists of a series of residual blocks with intermediate down-sampling layers and it embeds the given input image into a face embedding of size $h$.

\subsubsection{Audio Encoder.} The audio encoder is a standard CNN that takes a Mel-frequency cepstral coefficient (MFCC) heatmap of size $M$x$T$x$1$ and creates an audio embedding of size $h$. The audio embedding is concatenated with the face embedding to produce a joint audio-visual embedding of size $2$x$h$.

\subsubsection{Face Decoder.} This branch produces a lip-synchronized face from the joint audio-visual embedding by inpainting the masked region of the input image with an appropriate mouth shape. It contains a series of residual blocks with a few intermediate deconvolutional layers that upsample the feature maps. The output layer of the Face decoder is a sigmoid activated $1$x$1$ convolutional layer with $3$ filters, resulting in a face image of $H$x$H$x$3$. While ~\citet{chung2017you} employ only $2$ skip connections between the face encoder and the decoder, we employ $6$ skip connections, one after every upsampling operation to ensure that the fine-grained input facial features are preserved by the decoder while generating the face. As we feed the desired pose as input during training, the model generates a morphed mouth shape that matches the given pose. Indeed, in our results, it can be seen that we preserve the face pose and expression better than ~\citet{chung2017you} and only change the mouth shape. This allows us to seamlessly paste the generated face crop into the given video without any artefacts, which was not possible with ~\citet{chung2017you} due to the random uncontrollable pose variations.

\subsection{Discriminator network} While using only an L2 reconstruction loss for the generator can generate satisfactory talking faces~\cite{chung2017you}, employing strong additional supervision can help the generator learn robust, accurate phoneme-viseme mappings and also make the facial movements more natural. ~\citet{zhou2018talking} employed audio-visual speech recognition as a probe task to associate the acoustic and visual information. However, this makes the setup language-specific and offers only indirect supervision. We argue that directly testing whether the generated face synchronizes with the audio provides a stronger supervisory signal to the generator network. Accordingly, we create a network that encodes an input face and audio into fixed representations and computes the L2 distance $d$ between them. The face encoder and audio encoder are the same as used in the generator network.

\subsection{Joint training of the GAN framework} 

Our training process is as follows. We randomly select a $T$ millisecond window from an input video sample and extract its corresponding audio segment $A$, resampled at a frequency $F$~Hz. We choose the middle video frame $S$ in this window as the desired ground-truth. We mask the mouth region (assumed to be the lower-half of the image) of a person in the ground truth frame to get $S_m$. We also sample a negative frame $S'$, i.e., a frame outside this window which is expected to not be in sync with the chosen audio segment $A$. At each training batch to the generator, the unsynced face $S'$ concatenated channel wise with the masked ground truth face $S_m$ and the target audio segment $A$ is provided as the input. The generator is expected to generate the synced face $G([S'; S_m], A) \approx S$. Each training batch to the discriminator contains three types of samples: $(i)$ Synthetic samples from the generator $(G(S', A), A); y_i = 1, (ii)$ Actual frames synced with audio $(S, A); y_i = 0$ and $(iii)$ Actual frames out of sync with audio $(S', A); y_i = 1$. The third sample type is particularly important to force the discriminator to take into account the lip synchronization factor while classifying a given input pair as real / synthetic. Without the third type of sample, the discriminator would simply be able to ignore the audio input and make its decision solely on the quality of the image. The discriminator learns to detect synchronization by minimizing the following contrastive loss: 
\begin{equation}
{L_c(d_i, y_i) = \frac{1}{2N} \sum^{N}_{i=1} (y_i \cdot {d_i}^2 + (1 - y_i) \cdot max(0, m - d_i)^2)}
\end{equation}
where $m$ is the margin, which we set to $2$. The generator learns to reconstruct the face image by minimizing the L1 reconstruction loss:
\begin{equation}
{L_{Re}(G) = \frac{1}{N} \sum^{N}_{i=1} ||S - G(S', A)||_1}
\end{equation}
We train the generator $G$ and discriminator $D$ using the following GAN objective function:
% \begin{equation}
% L_{a}(G, D) = \mathbb{E}_{z,A}[L_c(D(z, A), y)] + \mathbb{E}_{S',A}[L_c(D(G(S', A), A), y=1)]
% \end{equation}

\begin{align}
L_{\mathrm{real}} &= \mathbb{E}_{z,A}[L_c(D(z, A), y)] \\ 
L_{\mathrm{fake}} &= \mathbb{E}_{S',A}[L_c(D(G([S';S_m], A), A), y=1)] \\ 
L_{a}(G, D) &= L_{\mathrm{real}} + L_{\mathrm{fake}} 
\end{align}

where $z \in \{S, S'\}$. Here, $G$ tries to minimize $L_{a}$ and $L_{Re}$ and $D$ tries to maximize $L_{a}$. Thus, the final objective function is:
\begin{equation}
G^* = \arg \min_G \max_D L_{a}(G, D) + L_{Re}
\end{equation}

\subsection{Implementation Details} We use the LRS 2 dataset~\cite{afouras2018deep} which contains over $29$ hours of talking faces in the provided train split in the dataset. We train on four NVIDIA TITAN X GPUs with a batch size of $512$. We extract $13$ MFCC features from each audio segment $(T = 350, F=100)$ and discard the first feature similar to ~\citet{chung2017you}. We detect faces in our input frame using \texttt{dlib}~\cite{king2009dlib} and resize the face crops to $96$x$96$x$3$. We use the Adam~\cite{kingma2014adam} optimizer with an initial learning rate of $1e{-3}$ and train for about $20$ epochs.

\subsection{Results and Evaluation}
\begin{figure*}[h]
  \includegraphics[width=470px, height=193px]{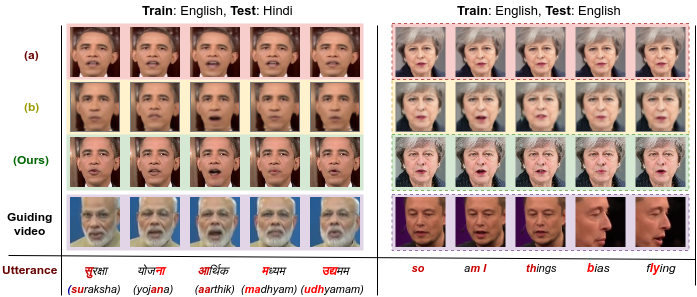}
  \caption{Visual comparison of faces generated by different models when they try to speak specific segments of the words shown in the last row. The audio segments corresponding to these word segments are extracted from the guiding video and fed into each of the models compared above. From top to bottom row: (a) ~\citet{zhou2018talking} (b) ~\citet{chung2017you} and Our LipGAN model. While (a) achieves poor lipsync across languages, and (b) generates unnatural lip movements, our LipGAN model produces consistent accurate, natural talking faces across languages.}
  %To compare the lip shapes We observed that (a) could not achieve lip sync across languages and a poor lipsync on words not seen during training. On the other hand, (b) achieves correct lip movements, but they look unnatural and lack clarity, which becomes even more evident when tested across languages (left half).}
  \label{fig:lip_synthesis}
\end{figure*}

We evaluate our novel LipGAN architecture quantitatively and also with subjective human evaluation. During inference, the model generates the talking face video of the target speaker frame-by-frame. The visual input is the current frame concatenated with the same current frame with the lower-half masked. That is, during inference, we expect the model to morph the input shape and preserve other aspects like pose and expression. Along with each of the visual inputs, we feed a $T = 350ms$ audio segment. In Figure \ref{fig:lip_synthesis}, we compare the talking faces generated by $3$ models on audio segments actually spoken by Narendra Modi and Elon Musk.

\subsubsection{Quantitative evaluation.}To evaluate our lip synthesis quantitatively, we use the LRW test set~\cite{chung2016lip}. We follow the same inference method mentioned above, but with one change. Instead of feeding the current frame as input as mentioned above, we feed a random input frame of the speaker, concatenated with the masked current frame for the pose prior. This is done to ensure we do not leak any lip information to the model while computing the quantitative metrics. In Table \ref{tab:lipgan_quant}, we report the scores obtained using standard metrics: PSNR, SSIM~\cite{wang2004image} and Landmark distance~\cite{chen2018lip}. As can be seen in Table \ref{tab:lipgan_quant}, our model significantly outperforms existing works across all quantitative metrics. These results highlight the superior quality of our generated faces (judged by PSNR) and also a highly accurate lip synthesis (LMD, SSIM). The noted increase in SSIM and the decrease in LMD can be attributed to the direct lip-synchronization supervision provided by the discriminator, which is absent in prior works.

\begin{table}[h]
  \begin{tabular}{l|cccc}
    \hline
    Algorithm & PSNR & SSIM & LMD \\
    \hline
    \citet{chung2017you}& 28.06 & 0.460 & 2.22\\
    \citet{zhou2018talking} & 26.80 & 0.884 & - \\
    \textbf{LipGAN (Ours)} & ~\textbf{33.4} & ~\textbf{0.960} & ~\textbf{0.60}\\
  \bottomrule
 \end{tabular}
  \caption{Our proposed LipGAN model achieves significant improvements over existing competitive approaches across all standard quantitative metrics.}
  \label{tab:lipgan_quant}
  \vspace{-0.7cm}
\end{table}

\subsubsection{Importance of the lip sync discriminator.} 
\begin{figure}[h]
  \includegraphics[width=230px, height=115px]{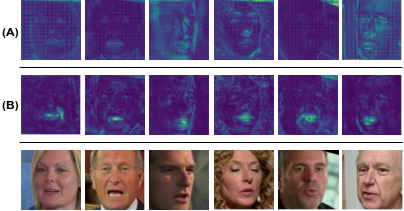}
  \caption{Activation heatmaps from the penultimate layer of two generator networks, one trained without a lip-sync discriminator (A) and the LipGAN network (ours) with a discriminator (B). Our network with the discriminator is highly attentive towards lip and mouth regions.}
  \label{fig:ablation}
  \vspace{-0.3cm}
\end{figure}

To illustrate the effect of employing a discriminator in the LipGAN network that tests whether the generator faces are in sync, we conduct the following experiment. We train the talking face generator network separately on the same train split of LRS 2 without changing any of the other hyperparameters. We feed the unseen test images shown in Figure \ref{fig:ablation} along with unseen audio segments as input to our LipGAN network and the plain generator network that was trained without the discriminator. We plot the activations of the penultimate layer of the generator in both these cases. From the heatmaps in Figure \ref{fig:ablation}, it is evident that our LipGAN network learns to attend strongly on the lip and mouth regions compared to the one that is not trained with a lip-sync discriminator. These findings also concur with the significant increase in the quantitative metrics as well as the natural movement of the lip regions in the generated faces.

\begin{table}[h]
  \begin{tabular}{lcc}
    \toprule
    Approach & Lip-sync rate & Realistic rate\\
    \midrule
    \citet{zhou2018talking} & 2.41 & 2.42\\
    \citet{chung2017you} &  2.95 & 3.10\\
    \textbf{Ours} & \textbf{3.68} & \textbf{3.73}\\
  \bottomrule
\end{tabular}
    \caption{LipGAN achieves significantly higher scores for both realistic rate and the extent of lip synchronization}
    \label{tab:lip_res}
    \vspace{-0.7cm}
\end{table}

% \begin{table}
%   \begin{tabularx}{\linewidth}{p{3cm}|XXXXX|XX}
%     \hline
%     Method & ASR & NMT & TTS & Voice Transfer & LipGAN & Semantic Consistency & Overall Experience\\
%     \hline
%     Translated Subtitles & \checkmark & \checkmark & $\times$ & $\times$ & $\times$ & \textbf{3.45} & 2.1\\
%     + Dubbing & \checkmark & \checkmark & \checkmark &  $\times$ & $\times$ & 3.22 & 2.21\\
%     + Personalization & \checkmark & \checkmark & \checkmark & \checkmark & $\times$ & 3.16 & 2.54\\
%     \textbf{+ LipSync} & \checkmark & \checkmark & \checkmark & \checkmark & \checkmark & 3.16 & \textbf{2.96}\\
%     \hline
%     Manual dubbing & - & - & - & - & -$\times$ & 4.79 & 4.18\\
%     \textbf{+ LipGAN} & - & - & - & - & \checkmark & \textbf{4.80} & \textbf{4.55}\\
%   \hline
% \end{tabularx}
% \caption{User ratings for different ways to consume cross-language multimedia content.}
% \label{tab:overallmos}
% \end{table}

\subsubsection{Human evaluation.}Talking face generation is primarily done for direct human consumption. Hence, alongside the quantitative metrics, we also subject it to human evaluation. We choose $10$ audio samples, with an equal number of English and Hindi speech videos. For each audio sample, we generate talking faces using three different models for $5$ popular identities to yield a total of $150$ samples. We compare the faces generated by three different models: $(i)$ ~\citet{chung2017you}, $(ii)$ ~\citet{zhou2018talking} and $(iii)$ Our LipGAN model. We conduct a user study with the help of $20$ participants who are asked to rate each of the videos on a scale of $1$ to $5$ based on the extent of lip synchronization and realistic nature. As shown in Table \ref{tab:lip_res}, our model obtains significantly higher scores compared to existing works.

\subsection{Evaluating the complete pipeline}
Finally, our Face-to-Face translation pipeline with all the components put together is evaluated based on its impact on the end-user experience. We choose $5$ famous identities and generate talking face videos in Hindi of Andrew Ng, Obama, Modi, Elon Musk and Chris Anderson using our complete pipeline. We do this by choosing short videos of each of the above speakers speaking in English. We use our ASR and NMT modules to recognize the speech in English and translate it to Hindi. We use our Hindi TTS model to obtain speech in Hindi. We convert this speech to the voices of each of the above speakers using our CycleGAN models. Using these final voices, we generate talking face videos using our LipGAN network. We compare these videos against videos with $(i)$ English speech and automatically translated subtitles $(ii)$ automatic dubbing to Hindi $(iii)$ automatic dubbing with voice transfer and $(iv)$ automatic dubbing with voice transfer + lip synchronization. Additionally, we also benchmark human performance for speech-to-speech translation: $(v)$ Manual dubbing and $(vi)$ Manual Dubbing + automatic lip synchronization. We ask the users to rate the videos on a scale of $1 - 5$ for two attributes. First one is "Semantic consistency" to check whether the automatic pipelines preserve the meaning of the original speech and the second attribute is the "Overall user experience", where the user considers factors such as the realistic nature of the talking face and his/her comfort level. The results of this study are reported in Table \ref{tab:overallmos}.

\begin{table}[h]
\begin{tabular}{l c c }
    \toprule
    Method    &  \makecell{Semantic \\ Consistency} & \makecell{Overall \\ Experience}\\
    \midrule
    Automatic Translated Subtitles &  \textbf{3.45}        & 2.10\\
    + Automatic Dubbing         &  3.22                 & 2.21\\
    + Automatic Voice Transfer   &  3.16                 & 2.54\\
    \textbf{+ lip-sync}   &  3.16                 & \textbf{2.96}\\
    \midrule
   Manual dubbing       &  4.79                 & 4.18\\
    \textbf{+ lip-sync}    &  \textbf{4.80}        & \textbf{4.55}\\
    \bottomrule
\end{tabular}
\caption{User ratings for different ways to consume cross-language multimedia content.}
\label{tab:overallmos}
\vspace{-0.6cm}
\end{table}

The results present three major takeaways. Firstly, we observe that there are significant scopes for improvement in each of the modules of automatic speech-to-speech translation systems. Future improvements in each of the speech and text translation systems will improve the user study scores. Secondly, the increase in user scores by using lip synchronization after manual dubbing again validates the effectiveness of the LipGAN model. Finally, note that adding each of our automatic modules increases the user experience score, emphasizing the need for each of them. Our complete proposed system improves the overall user experience over traditional text-based and speech-based translation systems by a significant margin.

\section{Applications}
\label{section:applications}
Our face-to-face translation framework can be used in a lot of applications. The demo video available here\footnote{http://cvit.iiit.ac.in/research/projects/cvit-projects/facetoface-translation} demonstrates a proof-of-concept for each of these applications.

% \subsection{Video Calling}
% This system can be potentially useful for applications like video calling. As mentioned previously, video calling has gained a lot of popularity in recent years. Often users from distant regions prefer speaking in their local language. Thus, inevitably the need for a cross language video calling setup has been felt for long. The Skype translator \cite{} was launched in October, 2015 which had features like speech-to-speech translation. But due to the absence of speaker's lip synchronization with the translated audio the user experience is hampered. In addition to speech-to-speech translation our system also produces videos of an user speaking in the translated language. Though our system at its current setup cannot be adopted directly for video calling but it can be extended in future for such an application.  
\vspace{-5pt}
\subsection{Movie dubbing} Movies are generally dubbed by dubbing artists manually. The dubbed audio is then overlaid to the original video. This causes the actors' lips to be out of sync with the audio, thus affecting the viewer experience. Our pipeline can be used to automate this process at different levels with different trade-offs. We demonstrate that we can synthesize and synchronize lips in manually dubbed videos, thus automatically correcting any dubbing errors.

We also show a proof-of-concept for performing automatic dubbing using our translation pipeline. That is, given a movie scene in a particular language our system can potentially be used to dub it to a different language. However, as shown by the user study scores in Table \ref{tab:overallmos}, significant improvements to the speech-to-speech pipeline is necessary to achieve realistic dubbing of complex speech present in movies. 

\subsection{Educational videos} As reported before in \cite{crosslanguage}, a large amount of online educational content is present in English in the form of video lectures. They are often aided with subtitles of foreign languages. But this increases the cognitive load of the viewer. Dubbing these videos with just speech-to-speech systems creates a visual discrepancy between the lip motion and the dubbed audio. However, with the help of our face to face translation system, it is possible to automatically translate such educational video content and also ensure lip synchronization.

%\subsection{Automatic Computer Facial Animation}
%Animating computer generated faces requires a skilled person to carefully observe the speech audio and manually overlay visemes on the animated faces. Such practices can be expensive and time-consuming. In our demo video, we show that it is possible to animate cartoon faces with LipGAN (despite not being trained on cartoon faces) using only audio inputs.

\subsection{Television news and interviews}
Automatic Face-to-Face Translation systems can potentially allow viewers to access and consume important information from across the globe irrespective of the underlying language. For example, a Hindi or German viewer can watch an English interview of Obama in the language of his/her choice with lip synchronization.

\section{Conclusion}
\label{section:conclusion}
We extend the problem of automatic machine translation to face to face translation with a focus on audio-visual content, i.e., where input and output are talking face videos. Beyond demonstrating the feasibility of a Face-to-Face translation pipeline, we also introduce a novel approach for talking face generation. We also contribute towards several language processing tasks (such as textual machine translation) for resource-constrained languages. Finally, we manifest our work on practical applications such as automatically dubbing educational videos, movie clips and interviews. 
Our attempt of "Face-to-Face Translation" also opens up a number of research directions in computer vision, multimedia processing, and machine learning. For instance, the duration of the speech gets naturally modified upon translation. This demands the transformation of the corresponding gestures, expressions, and background content. In addition to improving the existing individual modules, we believe that the above directions are all open for exploration.

\newpage
\bibliographystyle{ACM-Reference-Format}
\balance
\bibliography{references.bib}


\begin{thebibliography}{35}

%%% the Web address.

\ifx \showCODEN    \undefined \def \showCODEN     #1{\unskip}     \fi
\ifx \showDOI      \undefined \def \showDOI       #1{#1}\fi
\ifx \showISBNx    \undefined \def \showISBNx     #1{\unskip}     \fi
\ifx \showISBNxiii \undefined \def \showISBNxiii  #1{\unskip}     \fi
\ifx \showISSN     \undefined \def \showISSN      #1{\unskip}     \fi
\ifx \showLCCN     \undefined \def \showLCCN      #1{\unskip}     \fi
\ifx \shownote     \undefined \def \shownote      #1{#1}          \fi
\ifx \showarticletitle \undefined \def \showarticletitle #1{#1}   \fi
\ifx \showURL      \undefined \def \showURL       {\relax}        \fi
% invisible to TeX
\providecommand\bibfield[2]{#2}
\providecommand\bibinfo[2]{#2}
\providecommand\natexlab[1]{#1}
\providecommand\showeprint[2][]{arXiv:#2}

\bibitem[\protect\citeauthoryear{Afouras, Chung, Senior, Vinyals, and
  Zisserman}{Afouras et~al\mbox{.}}{2018}]%
        {afouras2018deep}
\bibfield{author}{\bibinfo{person}{Triantafyllos Afouras},
  \bibinfo{person}{Joon~Son Chung}, \bibinfo{person}{Andrew Senior},
  \bibinfo{person}{Oriol Vinyals}, {and} \bibinfo{person}{Andrew Zisserman}.}
  \bibinfo{year}{2018}\natexlab{}.
\newblock \showarticletitle{Deep audio-visual speech recognition}.
\newblock \bibinfo{journal}{\emph{IEEE transactions on pattern analysis and
  machine intelligence}} (\bibinfo{year}{2018}).
\newblock


\bibitem[\protect\citeauthoryear{Aharoni, Johnson, and Firat}{Aharoni
  et~al\mbox{.}}{2019}]%
        {aharoni2019massively}
\bibfield{author}{\bibinfo{person}{Roee Aharoni}, \bibinfo{person}{Melvin
  Johnson}, {and} \bibinfo{person}{Orhan Firat}.}
  \bibinfo{year}{2019}\natexlab{}.
\newblock \showarticletitle{Massively Multilingual Neural Machine Translation}.
\newblock \bibinfo{journal}{\emph{arXiv preprint arXiv:1903.00089}}
  (\bibinfo{year}{2019}).
\newblock


\bibitem[\protect\citeauthoryear{Amodei, Ananthanarayanan, Anubhai, Bai,
  Battenberg, Case, Casper, Catanzaro, Cheng, Chen, et~al\mbox{.}}{Amodei
  et~al\mbox{.}}{2016}]%
        {amodei2016deep}
\bibfield{author}{\bibinfo{person}{Dario Amodei}, \bibinfo{person}{Sundaram
  Ananthanarayanan}, \bibinfo{person}{Rishita Anubhai},
  \bibinfo{person}{Jingliang Bai}, \bibinfo{person}{Eric Battenberg},
  \bibinfo{person}{Carl Case}, \bibinfo{person}{Jared Casper},
  \bibinfo{person}{Bryan Catanzaro}, \bibinfo{person}{Qiang Cheng},
  \bibinfo{person}{Guoliang Chen}, {et~al\mbox{.}}}
  \bibinfo{year}{2016}\natexlab{}.
\newblock \showarticletitle{Deep speech 2: End-to-end speech recognition in
  english and mandarin}. In \bibinfo{booktitle}{\emph{International conference
  on machine learning}}. \bibinfo{pages}{173--182}.
\newblock


\bibitem[\protect\citeauthoryear{Arik, Chen, Peng, Ping, and Zhou}{Arik
  et~al\mbox{.}}{2018}]%
        {arik2018neural}
\bibfield{author}{\bibinfo{person}{Sercan Arik}, \bibinfo{person}{Jitong Chen},
  \bibinfo{person}{Kainan Peng}, \bibinfo{person}{Wei Ping}, {and}
  \bibinfo{person}{Yanqi Zhou}.} \bibinfo{year}{2018}\natexlab{}.
\newblock \showarticletitle{Neural voice cloning with a few samples}. In
  \bibinfo{booktitle}{\emph{Advances in Neural Information Processing
  Systems}}. \bibinfo{pages}{10040--10050}.
\newblock


\bibitem[\protect\citeauthoryear{Bahdanau, Cho, and Bengio}{Bahdanau
  et~al\mbox{.}}{2014}]%
        {bahdanau2014neural}
\bibfield{author}{\bibinfo{person}{Dzmitry Bahdanau},
  \bibinfo{person}{Kyunghyun Cho}, {and} \bibinfo{person}{Yoshua Bengio}.}
  \bibinfo{year}{2014}\natexlab{}.
\newblock \showarticletitle{Neural machine translation by jointly learning to
  align and translate}.
\newblock \bibinfo{journal}{\emph{arXiv preprint arXiv:1409.0473}}
  (\bibinfo{year}{2014}).
\newblock


\bibitem[\protect\citeauthoryear{Bregler, Covell, and Slaney}{Bregler
  et~al\mbox{.}}{1997}]%
        {bregler1997video}
\bibfield{author}{\bibinfo{person}{Christoph Bregler}, \bibinfo{person}{Michele
  Covell}, {and} \bibinfo{person}{Malcolm Slaney}.}
  \bibinfo{year}{1997}\natexlab{}.
\newblock \showarticletitle{Video Rewrite: driving visual speech with audio.}.
  In \bibinfo{booktitle}{\emph{Siggraph}}, Vol.~\bibinfo{volume}{97}.
  \bibinfo{pages}{353--360}.
\newblock


\bibitem[\protect\citeauthoryear{Chen, Li, K~Maddox, Duan, and Xu}{Chen
  et~al\mbox{.}}{2018}]%
        {chen2018lip}
\bibfield{author}{\bibinfo{person}{Lele Chen}, \bibinfo{person}{Zhiheng Li},
  \bibinfo{person}{Ross K~Maddox}, \bibinfo{person}{Zhiyao Duan}, {and}
  \bibinfo{person}{Chenliang Xu}.} \bibinfo{year}{2018}\natexlab{}.
\newblock \showarticletitle{Lip movements generation at a glance}. In
  \bibinfo{booktitle}{\emph{Proceedings of the European Conference on Computer
  Vision (ECCV)}}. \bibinfo{pages}{520--535}.
\newblock


\bibitem[\protect\citeauthoryear{Chung, Jamaludin, and Zisserman}{Chung
  et~al\mbox{.}}{2017}]%
        {chung2017you}
\bibfield{author}{\bibinfo{person}{Joon~Son Chung}, \bibinfo{person}{Amir
  Jamaludin}, {and} \bibinfo{person}{Andrew Zisserman}.}
  \bibinfo{year}{2017}\natexlab{}.
\newblock \showarticletitle{You said that?}
\newblock \bibinfo{journal}{\emph{arXiv preprint arXiv:1705.02966}}
  (\bibinfo{year}{2017}).
\newblock


\bibitem[\protect\citeauthoryear{Chung and Zisserman}{Chung and
  Zisserman}{2016}]%
        {chung2016lip}
\bibfield{author}{\bibinfo{person}{Joon~Son Chung} {and}
  \bibinfo{person}{Andrew Zisserman}.} \bibinfo{year}{2016}\natexlab{}.
\newblock \showarticletitle{Lip reading in the wild}. In
  \bibinfo{booktitle}{\emph{Asian Conference on Computer Vision}}. Springer,
  \bibinfo{pages}{87--103}.
\newblock


\bibitem[\protect\citeauthoryear{Federmann and Lewis}{Federmann and
  Lewis}{2016}]%
        {federmann2016microsoft}
\bibfield{author}{\bibinfo{person}{Christian Federmann} {and}
  \bibinfo{person}{William~D Lewis}.} \bibinfo{year}{2016}\natexlab{}.
\newblock \showarticletitle{Microsoft speech language translation (mslt)
  corpus: The iwslt 2016 release for english, french and german}. In
  \bibinfo{booktitle}{\emph{International Workshop on Spoken Language
  Translation}}.
\newblock


\bibitem[\protect\citeauthoryear{Griffin and Lim}{Griffin and Lim}{1984}]%
        {griffin1984signal}
\bibfield{author}{\bibinfo{person}{Daniel Griffin} {and} \bibinfo{person}{Jae
  Lim}.} \bibinfo{year}{1984}\natexlab{}.
\newblock \showarticletitle{Signal estimation from modified short-time Fourier
  transform}.
\newblock \bibinfo{journal}{\emph{IEEE Transactions on Acoustics, Speech, and
  Signal Processing}} \bibinfo{volume}{32}, \bibinfo{number}{2}
  (\bibinfo{year}{1984}), \bibinfo{pages}{236--243}.
\newblock


\bibitem[\protect\citeauthoryear{Jha, Namboodiri, and Jawahar}{Jha
  et~al\mbox{.}}{2019}]%
        {crosslanguage}
\bibfield{author}{\bibinfo{person}{Abhishek Jha}, \bibinfo{person}{Vinay
  Namboodiri}, {and} \bibinfo{person}{C~V Jawahar}.}
  \bibinfo{year}{2019}\natexlab{}.
\newblock \bibinfo{title}{Cross-Language Speech Dependent Lip-Synchronization}.
   (\bibinfo{year}{2019}).
\newblock
\newblock
\shownote{To appear in 2019 IEEE International Conference on Acoustics, Speech
  and Signal Processing (ICASSP).}


\bibitem[\protect\citeauthoryear{Johnson, Schuster, Le, Krikun, Wu, Chen,
  Thorat, Vi{\'e}gas, Wattenberg, Corrado, et~al\mbox{.}}{Johnson
  et~al\mbox{.}}{2017}]%
        {johnson2017google}
\bibfield{author}{\bibinfo{person}{Melvin Johnson}, \bibinfo{person}{Mike
  Schuster}, \bibinfo{person}{Quoc~V Le}, \bibinfo{person}{Maxim Krikun},
  \bibinfo{person}{Yonghui Wu}, \bibinfo{person}{Zhifeng Chen},
  \bibinfo{person}{Nikhil Thorat}, \bibinfo{person}{Fernanda Vi{\'e}gas},
  \bibinfo{person}{Martin Wattenberg}, \bibinfo{person}{Greg Corrado},
  {et~al\mbox{.}}} \bibinfo{year}{2017}\natexlab{}.
\newblock \showarticletitle{Google’s multilingual neural machine translation
  system: Enabling zero-shot translation}.
\newblock \bibinfo{journal}{\emph{Transactions of the Association for
  Computational Linguistics}}  \bibinfo{volume}{5} (\bibinfo{year}{2017}),
  \bibinfo{pages}{339--351}.
\newblock


\bibitem[\protect\citeauthoryear{Kaneko and Kameoka}{Kaneko and
  Kameoka}{2017}]%
        {kaneko2017parallel}
\bibfield{author}{\bibinfo{person}{Takuhiro Kaneko} {and}
  \bibinfo{person}{Hirokazu Kameoka}.} \bibinfo{year}{2017}\natexlab{}.
\newblock \showarticletitle{Parallel-data-free voice conversion using
  cycle-consistent adversarial networks}.
\newblock \bibinfo{journal}{\emph{arXiv preprint arXiv:1711.11293}}
  (\bibinfo{year}{2017}).
\newblock


\bibitem[\protect\citeauthoryear{King}{King}{2009}]%
        {king2009dlib}
\bibfield{author}{\bibinfo{person}{Davis~E King}.}
  \bibinfo{year}{2009}\natexlab{}.
\newblock \showarticletitle{Dlib-ml: A machine learning toolkit}.
\newblock \bibinfo{journal}{\emph{Journal of Machine Learning Research}}
  \bibinfo{volume}{10}, \bibinfo{number}{Jul} (\bibinfo{year}{2009}),
  \bibinfo{pages}{1755--1758}.
\newblock


\bibitem[\protect\citeauthoryear{Kingma and Ba}{Kingma and Ba}{2014}]%
        {kingma2014adam}
\bibfield{author}{\bibinfo{person}{Diederik~P Kingma} {and}
  \bibinfo{person}{Jimmy Ba}.} \bibinfo{year}{2014}\natexlab{}.
\newblock \showarticletitle{Adam: A method for stochastic optimization}.
\newblock \bibinfo{journal}{\emph{arXiv preprint arXiv:1412.6980}}
  (\bibinfo{year}{2014}).
\newblock


\bibitem[\protect\citeauthoryear{Kumar, Sotelo, Kumar, de~Br{\'e}bisson, and
  Bengio}{Kumar et~al\mbox{.}}{2017}]%
        {kumar2017obamanet}
\bibfield{author}{\bibinfo{person}{Rithesh Kumar}, \bibinfo{person}{Jose
  Sotelo}, \bibinfo{person}{Kundan Kumar}, \bibinfo{person}{Alexandre de
  Br{\'e}bisson}, {and} \bibinfo{person}{Yoshua Bengio}.}
  \bibinfo{year}{2017}\natexlab{}.
\newblock \showarticletitle{Obamanet: Photo-realistic lip-sync from text}.
\newblock \bibinfo{journal}{\emph{arXiv preprint arXiv:1801.01442}}
  (\bibinfo{year}{2017}).
\newblock


\bibitem[\protect\citeauthoryear{Kunchukuttan, Mehta, and
  Bhattacharyya}{Kunchukuttan et~al\mbox{.}}{2018}]%
        {kunchukuttan2018iit}
\bibfield{author}{\bibinfo{person}{Anoop Kunchukuttan}, \bibinfo{person}{Pratik
  Mehta}, {and} \bibinfo{person}{Pushpak Bhattacharyya}.}
  \bibinfo{year}{2018}\natexlab{}.
\newblock \showarticletitle{The IIT Bombay English-Hindi Parallel Corpus}. In
  \bibinfo{booktitle}{\emph{Proceedings of the Eleventh International
  Conference on Language Resources and Evaluation (LREC-2018)}}.
\newblock


\bibitem[\protect\citeauthoryear{Lewis}{Lewis}{2015}]%
        {skype}
\bibfield{author}{\bibinfo{person}{Will Lewis}.}
  \bibinfo{year}{2015}\natexlab{}.
\newblock \showarticletitle{Skype Translator: Breaking Down Language and
  Hearing Barriers}. In \bibinfo{booktitle}{\emph{Proceedings of Translating
  and the Computer (TC37)}}.
\newblock
\urldef\tempurl%
\url{https://www.microsoft.com/en-us/research/publication/skype-translator-breaking-down-language-and-hearing-barriers/}
\showURL{%
\tempurl}


\bibitem[\protect\citeauthoryear{Luong, Pham, and Manning}{Luong
  et~al\mbox{.}}{2015}]%
        {luong2015effective}
\bibfield{author}{\bibinfo{person}{Minh-Thang Luong}, \bibinfo{person}{Hieu
  Pham}, {and} \bibinfo{person}{Christopher~D Manning}.}
  \bibinfo{year}{2015}\natexlab{}.
\newblock \showarticletitle{Effective approaches to attention-based neural
  machine translation}.
\newblock \bibinfo{journal}{\emph{arXiv preprint arXiv:1508.04025}}
  (\bibinfo{year}{2015}).
\newblock


\bibitem[\protect\citeauthoryear{Neubig and Hu}{Neubig and Hu}{2018}]%
        {neubig2018rapid}
\bibfield{author}{\bibinfo{person}{Graham Neubig} {and} \bibinfo{person}{Junjie
  Hu}.} \bibinfo{year}{2018}\natexlab{}.
\newblock \showarticletitle{Rapid Adaptation of Neural Machine Translation to
  New Languages}. In \bibinfo{booktitle}{\emph{Proceedings of the 2018
  Conference on Empirical Methods in Natural Language Processing}}.
  \bibinfo{pages}{875--880}.
\newblock


\bibitem[\protect\citeauthoryear{NPD}{NPD}{2016}]%
        {videocalling}
\bibfield{author}{\bibinfo{person}{NPD}.} \bibinfo{year}{2016}\natexlab{}.
\newblock \bibinfo{title}{52 Percent of Millennial Smartphone Owners Use their
  Device for Video Calling, According to The NPD Group}.
\newblock
\newblock
\urldef\tempurl%
\url{https://www.npd.com/wps/portal/npd/us/news/press-releases/2016/52-percent-of-millennial-smartphone-owners-use-their-device-for-video-calling-according-to-the-npd-group/}
\showURL{%
\tempurl}


\bibitem[\protect\citeauthoryear{Panayotov, Chen, Povey, and
  Khudanpur}{Panayotov et~al\mbox{.}}{2015}]%
        {panayotov2015librispeech}
\bibfield{author}{\bibinfo{person}{Vassil Panayotov}, \bibinfo{person}{Guoguo
  Chen}, \bibinfo{person}{Daniel Povey}, {and} \bibinfo{person}{Sanjeev
  Khudanpur}.} \bibinfo{year}{2015}\natexlab{}.
\newblock \showarticletitle{Librispeech: an ASR corpus based on public domain
  audio books}. In \bibinfo{booktitle}{\emph{2015 IEEE International Conference
  on Acoustics, Speech and Signal Processing (ICASSP)}}. IEEE,
  \bibinfo{pages}{5206--5210}.
\newblock


\bibitem[\protect\citeauthoryear{Philip, Namboodiri, and Jawahar}{Philip
  et~al\mbox{.}}{2019}]%
        {philip2019baseline}
\bibfield{author}{\bibinfo{person}{Jerin Philip}, \bibinfo{person}{Vinay~P
  Namboodiri}, {and} \bibinfo{person}{CV Jawahar}.}
  \bibinfo{year}{2019}\natexlab{}.
\newblock \showarticletitle{A Baseline Neural Machine Translation System for
  Indian Languages}.
\newblock \bibinfo{journal}{\emph{arXiv preprint arXiv:1907.12437}}
  (\bibinfo{year}{2019}).
\newblock


\bibitem[\protect\citeauthoryear{Ping, Peng, Gibiansky, Arik, Kannan, Narang,
  Raiman, and Miller}{Ping et~al\mbox{.}}{2017}]%
        {ping2017deep}
\bibfield{author}{\bibinfo{person}{Wei Ping}, \bibinfo{person}{Kainan Peng},
  \bibinfo{person}{Andrew Gibiansky}, \bibinfo{person}{Sercan~O Arik},
  \bibinfo{person}{Ajay Kannan}, \bibinfo{person}{Sharan Narang},
  \bibinfo{person}{Jonathan Raiman}, {and} \bibinfo{person}{John Miller}.}
  \bibinfo{year}{2017}\natexlab{}.
\newblock \showarticletitle{Deep voice 3: Scaling text-to-speech with
  convolutional sequence learning}.
\newblock \bibinfo{journal}{\emph{arXiv preprint arXiv:1710.07654}}
  (\bibinfo{year}{2017}).
\newblock


\bibitem[\protect\citeauthoryear{Rousseau, Del{\'e}glise, and Esteve}{Rousseau
  et~al\mbox{.}}{2012}]%
        {rousseau2012ted}
\bibfield{author}{\bibinfo{person}{Anthony Rousseau}, \bibinfo{person}{Paul
  Del{\'e}glise}, {and} \bibinfo{person}{Yannick Esteve}.}
  \bibinfo{year}{2012}\natexlab{}.
\newblock \showarticletitle{TED-LIUM: an Automatic Speech Recognition dedicated
  corpus.}. In \bibinfo{booktitle}{\emph{LREC}}. \bibinfo{pages}{125--129}.
\newblock


\bibitem[\protect\citeauthoryear{Shen, Pang, Weiss, Schuster, Jaitly, Yang,
  Chen, Zhang, Wang, Skerrv-Ryan, et~al\mbox{.}}{Shen et~al\mbox{.}}{2018}]%
        {shen2018natural}
\bibfield{author}{\bibinfo{person}{Jonathan Shen}, \bibinfo{person}{Ruoming
  Pang}, \bibinfo{person}{Ron~J Weiss}, \bibinfo{person}{Mike Schuster},
  \bibinfo{person}{Navdeep Jaitly}, \bibinfo{person}{Zongheng Yang},
  \bibinfo{person}{Zhifeng Chen}, \bibinfo{person}{Yu Zhang},
  \bibinfo{person}{Yuxuan Wang}, \bibinfo{person}{Rj Skerrv-Ryan},
  {et~al\mbox{.}}} \bibinfo{year}{2018}\natexlab{}.
\newblock \showarticletitle{Natural tts synthesis by conditioning wavenet on
  mel spectrogram predictions}. In \bibinfo{booktitle}{\emph{2018 IEEE
  International Conference on Acoustics, Speech and Signal Processing
  (ICASSP)}}. IEEE, \bibinfo{pages}{4779--4783}.
\newblock


\bibitem[\protect\citeauthoryear{Sutskever, Vinyals, and Le}{Sutskever
  et~al\mbox{.}}{2014}]%
        {sutskever2014sequence}
\bibfield{author}{\bibinfo{person}{Ilya Sutskever}, \bibinfo{person}{Oriol
  Vinyals}, {and} \bibinfo{person}{Quoc~V Le}.}
  \bibinfo{year}{2014}\natexlab{}.
\newblock \showarticletitle{Sequence to sequence learning with neural
  networks}. In \bibinfo{booktitle}{\emph{Advances in neural information
  processing systems}}. \bibinfo{pages}{3104--3112}.
\newblock


\bibitem[\protect\citeauthoryear{Suwajanakorn, Seitz, and
  Kemelmacher-Shlizerman}{Suwajanakorn et~al\mbox{.}}{2017}]%
        {suwajanakorn2017synthesizing}
\bibfield{author}{\bibinfo{person}{Supasorn Suwajanakorn},
  \bibinfo{person}{Steven~M Seitz}, {and} \bibinfo{person}{Ira
  Kemelmacher-Shlizerman}.} \bibinfo{year}{2017}\natexlab{}.
\newblock \showarticletitle{Synthesizing obama: learning lip sync from audio}.
\newblock \bibinfo{journal}{\emph{ACM Transactions on Graphics (TOG)}}
  \bibinfo{volume}{36}, \bibinfo{number}{4} (\bibinfo{year}{2017}),
  \bibinfo{pages}{95}.
\newblock


\bibitem[\protect\citeauthoryear{Tachibana, Uenoyama, and Aihara}{Tachibana
  et~al\mbox{.}}{2018}]%
        {tachibana2018efficiently}
\bibfield{author}{\bibinfo{person}{Hideyuki Tachibana},
  \bibinfo{person}{Katsuya Uenoyama}, {and} \bibinfo{person}{Shunsuke Aihara}.}
  \bibinfo{year}{2018}\natexlab{}.
\newblock \showarticletitle{Efficiently trainable text-to-speech system based
  on deep convolutional networks with guided attention}. In
  \bibinfo{booktitle}{\emph{2018 IEEE International Conference on Acoustics,
  Speech and Signal Processing (ICASSP)}}. IEEE, \bibinfo{pages}{4784--4788}.
\newblock


\bibitem[\protect\citeauthoryear{Vaswani, Shazeer, Parmar, Uszkoreit, Jones,
  Gomez, Kaiser, and Polosukhin}{Vaswani et~al\mbox{.}}{2017}]%
        {vaswani2017attention}
\bibfield{author}{\bibinfo{person}{Ashish Vaswani}, \bibinfo{person}{Noam
  Shazeer}, \bibinfo{person}{Niki Parmar}, \bibinfo{person}{Jakob Uszkoreit},
  \bibinfo{person}{Llion Jones}, \bibinfo{person}{Aidan~N Gomez},
  \bibinfo{person}{{\L}ukasz Kaiser}, {and} \bibinfo{person}{Illia
  Polosukhin}.} \bibinfo{year}{2017}\natexlab{}.
\newblock \showarticletitle{Attention is all you need}. In
  \bibinfo{booktitle}{\emph{Advances in Neural Information Processing
  Systems}}. \bibinfo{pages}{5998--6008}.
\newblock


\bibitem[\protect\citeauthoryear{Wang, Bovik, Sheikh, Simoncelli,
  et~al\mbox{.}}{Wang et~al\mbox{.}}{2004}]%
        {wang2004image}
\bibfield{author}{\bibinfo{person}{Zhou Wang}, \bibinfo{person}{Alan~C Bovik},
  \bibinfo{person}{Hamid~R Sheikh}, \bibinfo{person}{Eero~P Simoncelli},
  {et~al\mbox{.}}} \bibinfo{year}{2004}\natexlab{}.
\newblock \showarticletitle{Image quality assessment: from error visibility to
  structural similarity}.
\newblock \bibinfo{journal}{\emph{IEEE transactions on image processing}}
  \bibinfo{volume}{13}, \bibinfo{number}{4} (\bibinfo{year}{2004}),
  \bibinfo{pages}{600--612}.
\newblock


\bibitem[\protect\citeauthoryear{Wu, Schuster, Chen, Le, Norouzi, Macherey,
  Krikun, Cao, Gao, Macherey, et~al\mbox{.}}{Wu et~al\mbox{.}}{2016}]%
        {wu2016google}
\bibfield{author}{\bibinfo{person}{Yonghui Wu}, \bibinfo{person}{Mike
  Schuster}, \bibinfo{person}{Zhifeng Chen}, \bibinfo{person}{Quoc~V Le},
  \bibinfo{person}{Mohammad Norouzi}, \bibinfo{person}{Wolfgang Macherey},
  \bibinfo{person}{Maxim Krikun}, \bibinfo{person}{Yuan Cao},
  \bibinfo{person}{Qin Gao}, \bibinfo{person}{Klaus Macherey}, {et~al\mbox{.}}}
  \bibinfo{year}{2016}\natexlab{}.
\newblock \showarticletitle{Google's neural machine translation system:
  Bridging the gap between human and machine translation}.
\newblock \bibinfo{journal}{\emph{arXiv preprint arXiv:1609.08144}}
  (\bibinfo{year}{2016}).
\newblock


\bibitem[\protect\citeauthoryear{Zen, Tokuda, and Black}{Zen
  et~al\mbox{.}}{2009}]%
        {zen2009statistical}
\bibfield{author}{\bibinfo{person}{Heiga Zen}, \bibinfo{person}{Keiichi
  Tokuda}, {and} \bibinfo{person}{Alan~W Black}.}
  \bibinfo{year}{2009}\natexlab{}.
\newblock \showarticletitle{Statistical parametric speech synthesis}.
\newblock \bibinfo{journal}{\emph{speech communication}} \bibinfo{volume}{51},
  \bibinfo{number}{11} (\bibinfo{year}{2009}), \bibinfo{pages}{1039--1064}.
\newblock


\bibitem[\protect\citeauthoryear{Zhou, Liu, Liu, Luo, and Wang}{Zhou
  et~al\mbox{.}}{2018}]%
        {zhou2018talking}
\bibfield{author}{\bibinfo{person}{Hang Zhou}, \bibinfo{person}{Yu Liu},
  \bibinfo{person}{Ziwei Liu}, \bibinfo{person}{Ping Luo}, {and}
  \bibinfo{person}{Xiaogang Wang}.} \bibinfo{year}{2018}\natexlab{}.
\newblock \showarticletitle{Talking Face Generation by Adversarially
  Disentangled Audio-Visual Representation}.
\newblock \bibinfo{journal}{\emph{arXiv preprint arXiv:1807.07860}}
  (\bibinfo{year}{2018}).
\newblock


\end{thebibliography}

\end{document}